\documentclass{article}
\usepackage[utf8]{inputenc}
\usepackage{graphicx}
\usepackage{amsmath, amssymb}
\usepackage{algorithm}
\usepackage{algorithmic}
\usepackage{hyperref}
\usepackage{bm}
\usepackage{booktabs}
\usepackage{float}
\usepackage{caption}
\usepackage{subcaption}
\usepackage{geometry}
\usepackage{soul,xcolor}
\usepackage{tikz}
\usepackage{bbm}
\usetikzlibrary{shapes.geometric, arrows, positioning, fit, calc}

\title{\vspace{-1.5cm}
Datarus-R1: An Adaptive Multi-Step Reasoning LLM for Automated Data Analysis}
\author{
    Ayoub Ben Chaliah \\
    \small \href{mailto:ayoub1benchaliah@gmail.com}{ayoub1benchaliah@gmail.com}
    \and
    Hela Dellagi \\
    \small \href{mailto:hela.dellagi@outlook.com}{hela.dellagi@outlook.com}
}
\date{\today} 

\begin{document}
\setstcolor{red}
\maketitle

\begin{abstract}
{\normalsize We present Datarus-R1-14B, a 14 B-parameter open-weights language model fine-tuned from Qwen 2.5-14B-Instruct to act as a virtual data analyst and graduate-level problem solver. Datarus is trained not on isolated question-answer pairs but on full analytical trajectories—including reasoning steps, code execution, error traces, self-corrections, and final conclusions— all captured in a ReAct-style notebook format spanning finance, medicine, numerical analysis, and other quantitative domains.
Our training pipeline combines (i) a trajectory-centric synthetic data generator that yielded 144 000 tagged notebook episodes, (ii) a dual-reward framework blending a lightweight tag-based structural signal with a Hierarchical Reward Model (HRM) that scores both single-step soundness and end-to-end coherence, and (iii) a memory-optimized implementation of Group Relative Policy Optimization (GRPO) featuring KV-cache reuse, sequential generation, and reference-model sharding. A cosine curriculum smoothly shifts emphasis from structural fidelity to semantic depth, reducing the format collapse and verbosity that often plague RL-aligned LLMs.
A central design choice in Datarus is it dual reasoning interface. In agentic mode the model produces ReAct-tagged steps that invoke Python tools to execute real code; in reflection mode it outputs compact Chain-of-Thought (CoT) traces delimited by \texttt{\small<think>} and \texttt{\small<answer>} tags. On demanding postgraduate-level problems, Datarus exhibits an “AHA-moment” pattern: it sketches hypotheses, revises them once or twice, and converges avoiding the circular, token-inflating loops common to contemporary systems.
Across standard public benchmarks Datarus surpasses similar size models and even reaches the level of larger reaoning models such as QwQ-32B achieving up to 30 \% higher accuracy on AIME 2024/2025 and LiveCodeBench while emitting 18–49 \% fewer tokens per solution. We release model weights and an interactive agentic pipeline for community use: 
\url{https://huggingface.co/DatarusAI/Datarus-R1-14B-preview},\url{https://github.com/DatarusAI/Datarus-JupyterAgent}.}
\end{abstract}

\begin{figure}[H]
    \centering
    \includegraphics[width=0.6\textwidth]{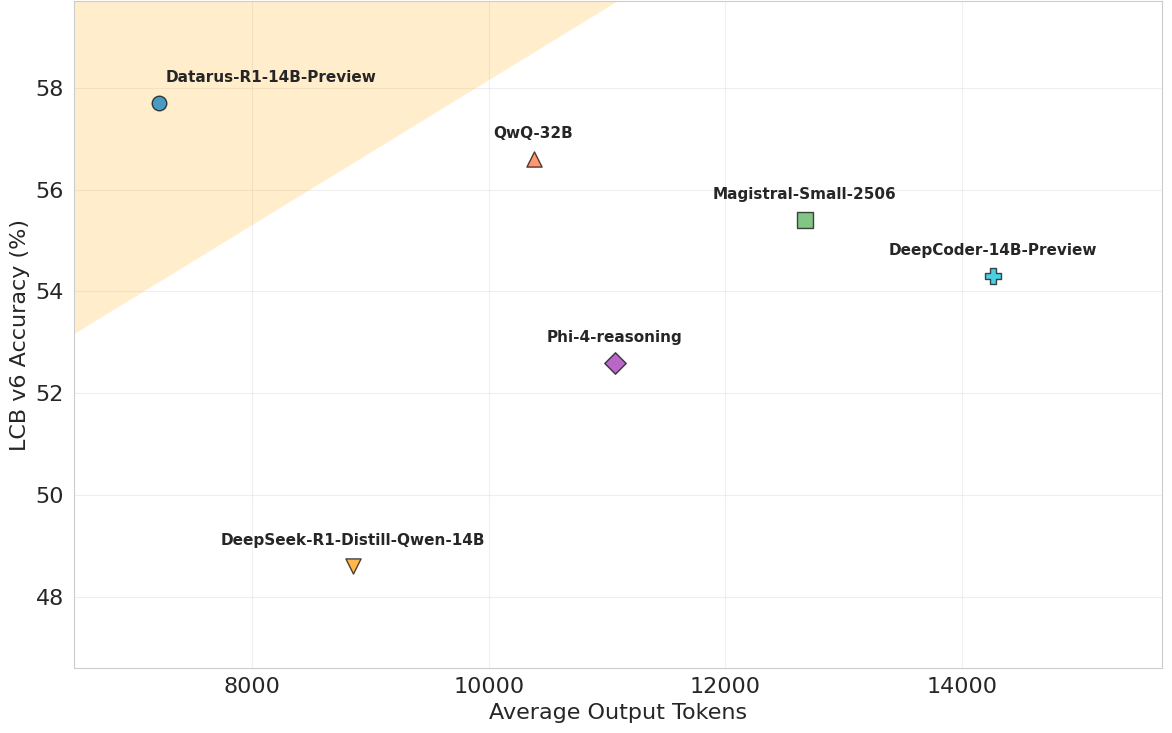}
    \caption{Performance-efficiency analysis on LCB v6 (2/24-4/25). Datarus-R1-14B-Preview sets a strong efficiency baseline for reasoning models, surpassing comparable and larger  models at significantly reduced token cost}
\end{figure}

\section{Introduction}

Large language models (LLMs) have transformed code generation, and mathematical reasoning through prompting techniques like Chain-of-Thought (CoT)~\cite{wei2022chain,openai2024learningreason} and ReAct~\cite{yao2022react}. Yet when facing real-world analytical challenges such as data analysis, formal proofs and intricate logic puzzles, most instruction-tuned models falter. They lack the \textbf{iterative, self-corrective workflows} that characterize expert analysis~\cite{hu2024infiagentdabench, zhang2024benchmarkingdsa}: formulating hypotheses, executing partial solutions, diagnosing errors, and refining their approach.\\
Practitioners engage in a loop of \textbf{reason → act → observe → reflect → revise}. Standard SFT datasets of static question-answer pairs fail to capture these rich trajectories, leading LLMs to either "overthink" with token-inflating loops or "underthink" by skipping key validation steps.\\
We argue that a truly capable analytical model must be \textbf{trained on the \textit{process} of reasoning}. Datarus-R1-14B realizes this through a \textbf{trajectory-centric} paradigm comprising four key innovations:

\begin{itemize}
\item \textbf{Trajectory-Centric Synthetic Data Generation:} 
  We derive domain-agnostic taxonomies from textbooks and technical sources across quantitative fields. We then use Qwen2.5-72B-Instruct~\cite{qwen-team2024qwen} to generate Python (or pseudo-code) scripts that create synthetic datasets embedding targeted challenges—ranging from numerical edge cases to combinatorial configurations and formal logic statements.  An agentic pipeline executes these scripts in a sandboxed notebook environment, capturing every \texttt{<step>}—thoughts, code cells, execution results (including errors), and reflections until a termination condition.  After human-in-the-loop validation, we obtain 144,000 high-quality trajectories stratified into four pedagogical categories that teach optimal solutions, error recovery, metacognitive correction, and avoidance of unproductive paths.

\item \textbf{Dual Reward Framework:} Our approach combines a \textbf{tag-based structural reward} that encourages clear, well-organized outputs by scoring early placement of \texttt{<step>} tags and rewarding semantic markers (\texttt{<thought>}, \texttt{<action\_input>}, \texttt{<stop\_analysis>}) and  a \textbf{Hierarchical Reward Model (HRM)}:  A Qwen2.5-3B network that assesses both individual steps and end-to-end trajectories, rewarding corrected mistakes and employing preference learning to teach \texttt{why} one reasoning path is superior to another

\item \textbf{Adaptive Curriculum Optimization:}
We refine Datarus with a curriculum that gradually shifts weight from structural to semantic rewards.  Early training phases enforce formatting habits, while later phases emphasize analytical correctness.  This approach prevents collapse of structure under semantic pressure and curbs unnecessary token overhead from overthinking.

\item \textbf{Dual Reasoning Interfaces:} 

\begin{itemize}
    \item \textbf{Agentic(ReAct) Mode:}  For interactive analysis, the model emits \texttt{<step>} blocks with embedded tool calls, enabling live code execution for data loading, simulation, symbolic manipulation, or visualization.
    \item \textbf{Reflection(CoT) Mode:} For concise documentation and proof-style expositions, the model produces \texttt{<think>}/\texttt{<answer>}  pairs that encapsulate complete reasoning chains in a compact format.    
\end{itemize}
\end{itemize}
In extensive evaluations, Datarus demonstrates a hallmark of expert cognition: \textbf{efficient hypothesis refinement}. On several challenging graduate-level problems the model formulates an initial solution approach, identifies and rectifies potential errors in a subsequent iteration, and arrives at the correct solution. This \textbf{concise revision cycle} of one to two iterations distinguishes Datarus from verbose models that engage in repetitive reasoning loops, consuming excessive tokens without corresponding improvements in solution quality. Across public benchmarks, Datarus-R1-14B-Preview \textbf{outperforms} open 14B models and \textbf{rivals} select 32B baselines on LiveCodeBench v6 (2/24 - 4/25) and AIME 2024/2025, with up to \textbf{30\% higher accuracy}. Our model consumes \textbf{ 18-49\% fewer tokens} per solution, delivering substantial inference cost savings, and \textbf{generalizes} seamlessly across diverse STEM challenges such as numerical analysis, combinatorial game theory, logic puzzles, and more, owing to its process-centric training. Its efficiency, accuracy, and dual-mode reasoning make Datarus well-suited to interactive notebooks, analytics pipelines, and automated reporting systems where both \textbf{explainability} and \textbf{compute control} are paramount.

\section{Background and Related Work}

The development of advanced reasoning models has converged on a powerful two-phase paradigm: Supervised Fine-Tuning (SFT) followed by Reinforcement Learning (RL)~\cite{chen2025acereason, guo2025deepseekr1, 2025magistral}. 
Recent breakthrough models like OpenAI's o1~\cite{openai2024learningreason} and DeepSeek-R1~\cite{guo2025deepseekr1} have demonstrated that extended reasoning chains can substantially improve performance on mathematical and coding tasks, establishing the SFT-then-RL paradigm as the foundation for state-of-the-art reasoning models. However, these models often suffer from "overthinking" generating unnecessarily verbose reasoning that inflates token costs without proportional accuracy gains~\cite{shrivastava2025samplemore, sui2025stopoverthinking}. \\
The design of reward signals has proven pivotal in reasoning model development, with early work on Process Reward Models (PRMs) demonstrating the value of step-by-step feedback~\cite{uesato2022solving, lightman2023letsverify}. Recent advances have explored hierarchical reward structures~\cite{wang2025hierarchicalreward} yet few have tackled the unique challenges of data analysis where correctness is nuanced and iterative refinement is essential.\\
Datarus handles these challenges through trajectory-centric training that explicitly models the iterative refinement process inherent to expert analysis, combined with: (i)data curation that filters high-overthinking traces, (ii) a curriculum-based GRPO~\cite{shao2024deepseekmath} that gradually tightens efficiency constraints by enforcing structural fidelity early and later, semantic precision providing a denser and more nuanced signal appropriate for iterative, open-ended task, (iii) and a dual reward framework that balances structural incentives for format consistency with a hierarchical reward model~\cite{wang2025hierarchicalreward} that ensures both local correctness and global coherence.
 The integration of language models with computational tools has emerged as a powerful paradigm, with ReAct~\cite{yao2022react} establishing structured reasoning-action cycles and subsequent work exploring agentic frameworks for specialized domains~\cite{luo2025largelanguage}. The adoption of a structured ReAct-style~\cite{yao2022react} format with explicit \texttt{<thought>}, \texttt{<action>}, and \texttt{<observation>} tags provides the cognitive scaffolding necessary for multi-step reasoning and tool use. While early work like Chain-of-Thought (CoT)~\cite{wei2022chain} elicited reasoning through prompting, Datarus internalizes this process through fine-tuning on structured trajectories of procedurally generated multi-domain problems that capture the complete methodological workflow of a data analyst. This integration of an agentic framework with a sophisticated, feedback-driven training methodology on the entire problem-solving journey including errors, dead ends, and self-corrections represents a significant step towards creating LLMs that can autonomously navigate the complexities of real-world data analysis~\cite{xie2023openagents, hong2024data}.

\section{Synthetic Data Generation}

We designed a multi-stage pipeline to generate high-quality, diverse, and realistic problem-solving trajectories that mirror real-world analytical challenges.  The complete dataset comprises 144,000 problem-solving trajectories derived from 20,000 unique datasets, each containing embedded analytical challenges of varying complexity.

\subsection{ Knowledge Distillation and Multi-Domain Challenge Design}
Our approach begins with systematic knowledge extraction from carefully curated document collections spanning finance, medicine, numerical analysis, biostatistics, economics, and related quantitative fields. We assembled comprehensive repositories of courses, textbooks, and technical documents for each target domain. Using Qwen2.5-72B-Instruct~\cite{qwen-team2024qwen} instruct, we implemented a multi-phase knowledge distillation process.

\begin{enumerate}
    \item \textbf{Hierarchical Taxonomy Construction: } We segment each document into chunks sized to fit within the model's context window. For each chunk, we prompt the model to extract main topics and subtopics in a structured format. This procedure generates fine-grained conceptual mappings of the examined domain, wherein overarching topics are systematically organized into corresponding subtopics (see table 1).
After processing all chunks within a document, we aggregate the extracted topics and subtopics, then prompt the model to refine and consolidate this collection into a coherent taxonomy. This refinement process reduces redundancies and ensures proper hierarchical organization. We perform sample based quality checks on randomly selected outputs to maintain accuracy.
Once all documents within a domain are processed, we perform a final domain-level refinement, iterating until we achieve satisfactory conceptual coverage and organization.
Along with the taxonomy extraction, we also construct structured schemas that formalize domain-specific analytical characteristics. These schemas consist of three key components:\\ (i) representative data fields that capture typical data elements relevant to the domain, (ii) key requirements that articulate essential analytical considerations, and (iii)analysis techniques that specify appropriate methodological approaches.\\
This methodology is applied systematically across all chosen domains, resulting in comprehensive knowledge maps that capture the breadth and depth of each field.

\item \textbf{Procedural Challenge Generation: }
    
Leveraging the domain taxonomies, we employ Qwen2.5-Coder-32B-Instruct~\cite{qwen-team2024qwencode} to generate Python scripts that create synthetic datasets containing field-specific challenges targeting sampled topics and subtopics of each domain.
The generation process draws from domain-specific requirements and analysis techniques defined in our structured metadata. For example, Financial challenges might involve detecting high-frequency trading dynamics within transaction flows, while medical challenges could focus on identifying biomarker patterns across patient monitoring trajectories. 
Generated scripts undergo execution in sandboxed environments with domain-specific validation protocols. Beyond ensuring successful execution and meaningful output through a suite of verification checks, 
 we verify that the embedded domain challenges are properly represented and detectable through the specified analytical techniques. Scripts failing validation checks and scripts producing execution errors undergo automated correction procedures before final retention.

\end{enumerate}

\begin{table}[h!]
\centering
\begin{tabular}{|l|c|c|}
\hline
\textbf{Domain} & \textbf{Extracted Topics} & \textbf{Extracted Subtopics} \\
\hline
Data Science        & 20 & 400 \\
Dynamic Systems     & 56 & 504 \\
Environment         & 35 & 419 \\
Finance             & 44 & 396 \\
Graph Theory        & 39 & 351 \\
Industry            & 60 & 540 \\
Linear Algebra      & 49 & 475 \\
Medicine            & 50 & 500 \\
Numerical Analysis  & 62 & 624 \\
Probability         & 54 & 457 \\
\hline
\end{tabular}
\caption{Hierarchical taxonomy extraction across domains: number of topics and subtopics identified per domain.}
\end{table}

\subsection{The Analyst Simulation Loop}
With synthetic datasets and their embedded challenges established, we implement a sophisticated simulation framework to generate realistic problem-solving trajectories.
The simulation framework operates through a structured execution-feedback loop designed to replicate authentic domain-specific problem solving workflows. This process generates comprehensive analytical trajectories that exhibit tree-like exploration patterns through Jupyter notebook-style execution sequences where each retry and correction attempt creates branching pathways. 
\begin{enumerate}
\item \textbf{Reasoning and Action:} The model is instructed to follow a ReAct-style format, generating text (Markdown) explaining its thought process, followed by a code cell to execute an action (e.g., load data, plot a distribution, run a statistical test).
\item \textbf{Agentic Execution:} The code cell is executed by our agentic pipeline. The output of the execution---whether it's a dataframe, a plot, a numerical result, or an error traceback is captured.

\item \textbf{Iterative Refinement:} The result (or error) is fed back to the analyst model, which then decides on the next step. This loop continues until one of three termination conditions is met:
\begin{itemize}
    \item \textbf{Persistent Error:} The model generates code that results in an error for several consecutive retries.
    \item \textbf{Maximum Steps:} The notebook reaches a predetermined maximum number of cells to prevent infinite loops.
    \item \textbf{Self-Termination:} After a variable minimum number of steps, the model is instructed to decide whether to continue or conclude the analysis by outputting a special \texttt{<stop\_analysis>} tag.
\end{itemize}
\end{enumerate}
The resulting notebook follows a structured format where each analysis step is organized using specific XML-like tags. This structure ensures consistency and enables automated parsing of the analysis trajectory:
\begin{itemize}
\item \texttt{<step>}: Encapsulates each complete analysis iteration, containing the reasoning, action, and observation components
\item \texttt{<thought>}: Contains the model's reasoning process and rationale for the next analytical step
\item \texttt{<action>}: Specifies the computational tool or method to be employed, in our case the python executor.
\item \texttt{<action\_input>}: Contains the actual Python code to be executed
\item \texttt{<observation>}: Captures and documents the execution results, including outputs, visualizations, or error messages
\item \texttt{<stop\_analysis>}: Signals the termination of the analysis loop when the model determines the investigation is complete
\item \texttt{<answer>}: Contains the final analysis conclusion and key findings
\end{itemize}
This structured approach enables systematic evaluation of the analyst's reasoning process, code quality, and ability to iteratively refine their analysis based on intermediate results.

\subsection{Trajectory Stratification}
This simulation process yields a rich dataset of problem-solving paths, which we define as trajectories. Human-in-the-loop sampling and verification were performed at each stage to ensure quality. The final 144,000 trajectories were stratified to provide diverse learning signals:

\begin{itemize}
    \item \textbf{Success Trajectories (40\%):}  Clean, first-attempt solutions that demonstrate optimal problem-solving paths. These teach efficiency and best practices.

    \item \textbf{Error-Correction Trajectories ( 35\%):} Notebooks that initially failed but were successfully corrected. These explicitly teach the model how to recognize and recover from specific errors. A portion of these were manually curated to create direct "bad answer" vs. "good answer" pairs.
    \item \textbf{Self-Correction Sequences (15\%):}   Trajectories where the model identifies and corrects its own logical or coding errors within the same reasoning turn. These are invaluable for teaching metacognitive awareness.

    \item \textbf{Persistent Failure Trajectories ( 10\%):} Approaches that consistently fail. These serve as negative examples, teaching the model to avoid unproductive solution paths.

\end{itemize}

\section{Training Methodology}
Datarus employs a two-phase training methodology consisting of Supervised Fine-Tuning (SFT) followed by Group Relative Policy Optimization (GRPO)~\cite{shao2024deepseekmath} . The SFT phase establishes structured reasoning capabilities using our synthetic trajectory dataset, while GRPO refines performance through our dual reward system that balances structural formatting with semantic correctness. The training was conducted on an 8 NVIDIA H200 GPUs.

\subsection{Supervised Fine-Tuning (SFT)}
We demonstrate that trajectory quality and structural consistency directly determine post-RL performance, with our carefully designed training data providing the foundation for subsequent reinforcement learning improvements~\cite{li2024gettingmorejuice}.
\subsubsection{Structured Reasoning Framework}
The SFT phase implements a carefully designed cognitive scaffolding through the ReAct (Reasoning-Action) paradigm adapted for data analysis. This structure serves multiple purposes beyond mere formatting:
It enforces explicit reasoning before action, preventing the common failure mode of premature implementation. The separation of thought, action, and observation creates natural checkpoints for evaluation and error recovery. The hierarchical structure enables granular credit assignment during training while maintaining human interpretability.

\subsubsection{ Data Processing Pipeline}
The notebook parsing system extracts structured information from each trajectory:
\begin{itemize}
    \item Scenario and objective metadata from initialization cells
    \item Step-by-step progression through reasoning, code execution, and reflection phases
    \item Complete execution outputs including both successes and failures
    \item Executive summaries synthesizing insights
\end{itemize}
This parsed structure undergoes conversion to a standardized ReAct format that preserves the logical flow while enforcing consistent structure across all examples.
The training data combines three sources in carefully calibrated proportions:
\begin{itemize}
    \item \textbf{Error-Free Trajectories (60\%)}  Error-free notebooks demonstrating successful problem-solving patterns. These examples teach domain-specific practices and the rhythm of effective data analysis.
    \item \textbf{Error-Correction Examples (20\%):}Merged trajectories showing the transition from failure to success. These examples use special markers to highlight corrective reasoning, teaching the model to recognize and respond to errors appropriately.
    \item \textbf{ Curated Reasonning Datasets Integration (20\%):} High-quality reasoning examples from diverse domains, filtered to remove overthinking patterns. This external data prevents overfitting to data analysis patterns while maintaining structural consistency.
\end{itemize}

\subsubsection{Overthinking Prevention}

\textbf{Identifying Overthinking Patterns:} We detect unproductive repetition in both ReAct and CoT modes by monitoring reuse of identical hypotheses in successive reasoning segments, recurring filler phrases like "let me check again," and long spans with minimal new information. These signals generate an overthinking score—trajectories exceeding thresholds are flagged or discarded.\\
\textbf{Dataset Curation:} We blend ReAct notebooks from synthetic data and curated CoT datasets. Individual traces are ranked by overthinking score, with sampling enforcing a 90/10 split: 90\% from lowest-scoring (concise, information-dense) examples and 10\% from deeper explorations with genuinely new reasoning steps. This prevents learning endless loops while preserving rich exploration capability.\\
\textbf{Reinforcement Regularization:} Our GRPO implementation includes a token-efficiency penalty for completions exceeding adaptive length horizons. The penalty weight increases over training, initially allowing exploration but progressively enforcing brevity.\\
\textbf{Results:} After filtering, average CoT chain length drops ~40\% without accuracy loss, median ReAct step count falls ~45\%, and manual inspection confirms the 10\% deep traces capture legitimate breakthroughs rather than circular restatements.

\subsection{GRPO with Dual Reward System}
\subsubsection{Architectural Innovation}

The GRPO implementation represents a non-trivial engineering effort in distributed training architecture.  We strategically partition computational resources to achieve a high training efficiency:

\vspace{0.2em}
\textbf{Advanced Training Infrastructure (GPUs 0-5):} We leverage DeepSpeed ZeRO Stage 3~\cite{rajbhandari2019zero}  optimization to enable seamless training of 14B+ parameter models. 
This achieves linear memory scaling without sacrificing computational efficiency through  sharding model states, optimizer states, and gradients across six GPUs. 
\vspace{0.2em}

\textbf{Dedicated Generation Service (GPUs 6-7):} We implemented a dedicated vLLM service that runs on separate GPUs, providing high-throughput generation for the multiple solution attempts required by GRPO. This separation is crucial—training and generation have conflicting memory access patterns, and co-location leads to memory fragmentation and out-of-memory errors.

\subsubsection{Memory-Optimized GRPO Implementation}

The GRPO algorithm required several key optimizations to scale to large models:
\vspace{0.2em}

\textbf{KV-Cache Reuse Strategy:} We perform prompt encoding per batch, with the resulting key-value pairs cached and reused across all generation attempts. This reduces redundant computation by 75\% for typical 4\-generation batches.
\vspace{0.2em}

\textbf{Sequential Generation Processing:} Rather than storing all generations in memory simultaneously, the system processes each generation sequentially, computing gradients and accumulating loss before moving to the next. We reduce peak memory usage by a factor of num\_generations through this intelligent approach.

\textbf{Reference Model Sharding:} The reference model, used for KL divergence computation, is sharded across the training GPUs using DeepSpeed's inference engine. This eliminates the need for a separate full model copy, saving 14B parameters worth of memory.
\vspace{0.2em}

\textbf{Numerical Stability Enhancements:} Our implementation includes critical patches for training stability:
\begin{itemize}
    \item Delta clamping (-4 to 4) before exponentiation prevents numerical overflow
    \item Off-by-one alignment fix ensures proper next-token prediction
    \item Gradient clipping at token level prevents explosion in long sequences
\end{itemize}

\subsubsection{Tag-Based Structural Reward Design}

Our tag-based reward system implements nuanced incentives that shape output structure without being overly prescriptive:

\vspace{0.5em}
\textbf{Distance-Based Scoring for \texttt{<step>}:}  We apply distance-based scoring exclusively to the primary structural tag \texttt{<step>}:
\begin{verbatim}
step_score = 1.0 - (position / text_length) if "<step>" in text else 0.0
\end{verbatim}
This creates a smooth gradient pulling the step tag toward the beginning, ensuring responses start with proper structure rather than preamble.

\vspace{0.5em}
\textbf{Presence Bonuses for Semantic Tags:} Other tags receive fixed bonuses for presence, not position:
\begin{verbatim}
tag_bonuses = {
    "<thought>": 0.8,      # Highest - reasoning is critical
    "<action>": 0.6,       # Tool use indication
    "<action_input>": 0.4, # Implementation details
    "</step>": 0.2,        # Proper closure
    "<stop_analysis>": 0.6 # Final analysis marker
}
\end{verbatim}
This design is intentional, forcing all tags to appear early would create unnatural, front-loaded responses. Instead, semantic tags can appear where logically appropriate while the \texttt{<step>} tag ensures immediate structure.

\vspace{0.5em}
\textbf{Compound Tag Recognition:} The system also recognizes tag sequences like \texttt{<stop\_analysis><thought>} and \texttt{<stop\_analysis><answer>}, rewarding proper conclusion formatting without dictating exact positions.

\subsubsection{Hierarchical Reward Model (HRM) Architecture}
Unlike mathematical problems with definitive correct answers or coding challenges with unit test verification, data analysis requires nuanced evaluation of reasoning quality, methodological appropriateness, and iterative problem-solving effectiveness.
We built our HRM~\cite{wang2025hierarchicalreward} on Qwen2.5-3B~\cite{qwen2024technical} to address the fundamental challenge of evaluating data analysis workflows where traditional binary verification is insufficient. It evaluates semantic correctness through multiple lenses:

\vspace{0.5em}
\textbf{Process-Level Evaluation:} Single-step rewards assess immediate correctness: did this code execute? Is the reasoning sound? This provides dense supervision signals.

\vspace{0.5em}
\textbf{Trajectory-Level Evaluation:} Multi-step sequences are evaluated holistically through our innovative labeling strategy:
\begin{itemize}
    \item Sequences ending in error: Negative reward
    \item Sequences with errors that are corrected: Positive reward
    \item Error-free sequences: Positive reward
\end{itemize}
This teaches the model that mistakes are acceptable if properly addressed—a crucial lesson for robust problem-solving.

\vspace{0.5em}
\textbf{Preference Learning:} Failed and successful trajectory pairs teach relative quality assessment. The model learns not just what's correct, but why one approach is superior to another.

\subsubsection{Dynamic Lambda Scheduling}
We implemented a cosine schedule for reward weighting that implements a staged curriculum:

\begin{align}
\lambda_{\text{tag}} &= \frac{1 + \cos(\pi \times \text{training\_step} / \text{total\_steps})}{2}\\
\lambda_{\text{hrm}} &= 1 - \lambda_{\text{tag}}
\end{align}

\vspace{0.5em}
\textbf{Early Phase ($\lambda_{\text{tag}} \approx 1.0$):} The model focuses on learning proper structure. Tag rewards dominate, establishing formatting habits without concern for semantic correctness.

\vspace{0.5em}
\textbf{Transition Phase:} As $\lambda_{\text{tag}}$ decreases and $\lambda_{\text{hrm}}$ increases, the model must maintain structure while improving solution quality. This prevents format collapse—a common issue where models abandon structure when semantic rewards dominate.

\vspace{0.5em}
\textbf{Final Phase ($\lambda_{\text{hrm}} \approx 1.0$):} With structure internalized, the model refines solution quality. The small residual tag reward prevents format degradation.

\subsubsection{Distributed Generation Pipeline}

The vLLM~\cite{kwon2023pagedattention} integration required careful engineering to prevent bottlenecks through three key mechanisms. First, prompts are filtered before generation to ensure they fit within vLLM's context window with room for completions, preventing mid-generation failures that would desynchronize the distributed system.\\
To maintain synchronization across distributed ranks, a collective communication protocol coordinates the generation process. All ranks filter their prompts locally, with valid prompts gathering to rank 0. Rank 0 then communicates with the vLLM service, after which completions scatter back to originating ranks while empty completions fill skipped prompt slots. This protocol ensures all ranks remain synchronized even when some prompts are skipped.\\
The system supports zero-shot model updates through a request-response protocol that allows the vLLM service to load new checkpoints without restarting, ensuring no in-flight generations are lost during model updates.

\subsubsection{Emergent Training Phenomena}

The dual reward system creates rich learning dynamics with three distinct emergent patterns. Most notably, structural convergence follows a predictable trajectory as the \texttt{<step>} tag migration evolves from random placement with high variance (steps 0-50), through rapid migration toward beginning (steps 50-150), to final convergence on immediate placement (steps 150+).\\
Additionally, the model discovers natural tag partnerships through co-occurrence learning. \texttt{<thought>} almost always follows \texttt{<step>}, \texttt{<action>} triggers \texttt{<action\_input>}, and \texttt{<stop\_analysis>} pairs with \texttt{<answer>}. These patterns emerge organically from data statistics rather than hard-coded rules.\\
By step 100+ the model develops sophisticated error signature recognition, building an implicit catalog of debugging patterns. KeyError prompts checking column names, ValueError indicates examining data types, and IndexError requires verifying data dimensions. This accumulated knowledge manifests as increasingly targeted debugging strategies.

\subsubsection{Production-Scale Optimizations}

Several complementary optimizations enable stable training at scale through memory and computational efficiency improvements. The system employs gradient accumulation with batch size 1 and 16 accumulation steps to achieve an effective batch size of 16 while minimizing memory usage. Simultaneously, mixed precision training uses BF16 computation with FP32 master weights to balance numerical stability with memory efficiency.\\
Beyond memory optimizations, the system leverages computational trade-offs for scalability. Activation checkpointing enables larger models by trading computation for memory through selective recomputation during backward pass. Meanwhile, asynchronous checkpointing prevents training stalls by overlapping model saves with training through DeepSpeed's async I/O, ensuring continuous training progress.

\section{Evaluation and Results}
We evaluated Datarus-R1-14B-Preview against state-of-the-art reasoning models across comprehensive
benchmarks: LiveCodeBench~\cite{jain2024livecodebench} for code generation, AIME 2024 and 2025 for mathematical reasoning and  GPQA Diamond~\cite{rein2024gpqa} for scientific domain knowledge. All evaluations used greedy decoding except where noted, with 8 seeds for AIME evaluations to ensure statistical significance.\\Datarus demonstrates exceptional performance across all benchmarks, establishing new state-of-the-art results within the 14B parameter class. The model achieves superior performance on LiveCodeBench evaluations, surpassing comparable models while also competing with substantially larger models. In mathematical reasoning, Datarus-R1-14B-Preview exhibits robust capabilities with scores of 70.1 on AIME 2024 and 66.2 on AIME 2025, the latter representing the highest performance among 14B parameter models. The model's scientific reasoning proficiency is evidenced by its 62.1\% accuracy on GPQA Diamond, substantially outperforming other models both in and above its parameter class. \\
\begin{table}[h!]
\centering
\resizebox{\textwidth}{!}{%
\begin{tabular}{|l|c|c|c|c|c|c|c|}
\hline
\textbf{Model} & \textbf{Model Size} & \textbf{LCB v5 (8/24-1/25)} & \textbf{LCB v6 (2/24-4/25)} &  \textbf{LCB v6} & \textbf{AIME 24} & \textbf{AIME 25} & \textbf{GPQA D} \\
\hline
\textbf{Datarus-14B preview} & 14.8 B & \textbf{59.5} & \textbf{57.7} & \textbf{69.1} & 70.1 (4.8) & \textbf{66.2 (6.1)} & \textbf{62.1} \\
DeepCoder-14B (FP32) & 14.8 B & 59.1 & 54.3 & 67.5 & 63.7 (3.9) & 51.2 (7.6) & 55.0  \\
Phi-4-reasoning & 14.7 B & 58.4 & 52.6 & 67.5 & \textbf{74.6 (6.3)*} & 63.1 (6.3)* & 55.0  \\
DeepSeek-R1-Distill-14B & 14.8 B & 55.2 & 48.6 & 63.3 & - & - & 58.6  \\
\hline
Magistral-Small-2506 & 23.6 B & 66.3 & 55.4 & 68.25 & 70.7* & 62.7* & 56.6 \\
QwQ 32B & 32.8 B & 68.1 & 56.6 & 73.8 & 76.2 (4.2) & 66.2 (3.9) & 60.1 \\
DeepSeek-V3 & 685 B & 36.2* & - & - & 39.2* & - & 59.1* \\
DeepSeek-R1 & 685 B  & 65.9* & - & - & 79.8* & - & 71.5* \\
\hline
\end{tabular}}
\caption{Benchmark results across multiple models. Values in parentheses are standard deviations and * denotes reported values from official paper}
\end{table}
A key contribution of Datarus is its \textbf{token efficiency}. By adaptively modulating reasoning depth, the model avoids unnecessary verbose reasoning (``overthinking'') while still retaining sufficient depth for complex tasks.
On average, Datarus generates \textbf{18-49\% fewer tokens} per problem than competing models( see figure 2).
The efficiency gains are most dramatic against verbose models like DeepCoder-14B~\cite{luo2025deepcoder} (49\% reduction) and Magistral-Small-2506~\cite{2025magistral}  (43\% reduction), while still achieving meaningful savings against more efficient competitors like DeepSeek-R1-Distill-14B~\cite{guo2025deepseekr1}(18.5\% reduction) and QwQ-32B~\cite{qwen-team2025qwq} (30.5\% reduction).\\
The efficiency advantage grows with task difficulty: competing models exhibit dramatic token inflation on harder problems while Datarus maintains relatively modest token usage across all difficulty levels with competitive or superior performance. Most notably, Phi-4-reasoning~\cite{abdin2024phi4} explodes from 2159.8 Avg output tokens on easy problems to 20399.6 on hard problems, a staggering 945\% increase and  QwQ-32B shows a 600\% increase (from 3,031 to 18,192 Avg output tokens).
This efficiency directly translates into \textbf{lower inference cost} and \textbf{faster response times}, a critical factor for production deployment.\\
\begin{figure}[H]
    \centering
    \includegraphics[width=0.8\textwidth]{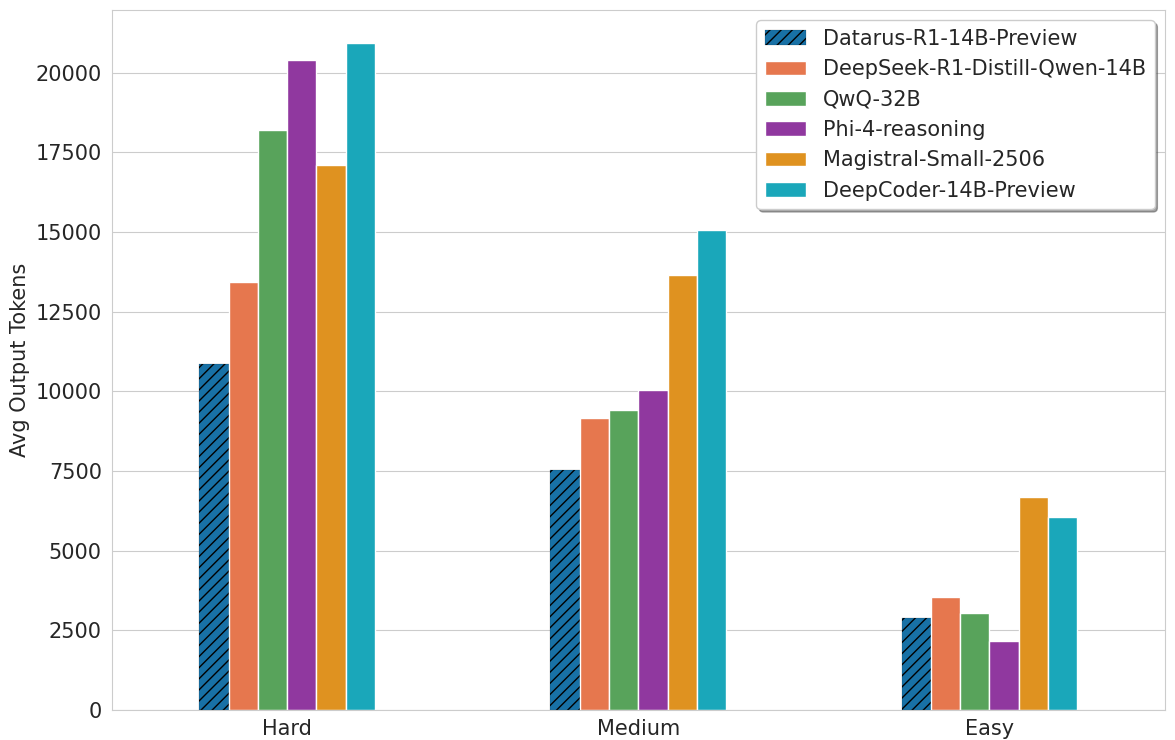}
    \caption{Comparative Analysis of Model Verbosity by Difficulty Level on LiveCodeBench v6 Tasks}
\end{figure}
Because Datarus-R1-14B-preview operates with \textbf{14B parameters in BF16 precision}, its combination of \textbf{parameter efficiency} and \textbf{adaptive reasoning} enables significantly higher throughput under limited compute budgets. 
This \emph{``no overthinking'' approach}---automatically calibrating reasoning depth to task complexity---marks an advancement in reasoning models' training design. It ensures that Datarus delivers \textbf{high-quality outputs at a fraction of the cost}, making it especially attractive for real-world deployments where both quality and efficiency are paramount.

\section{Conclusion }

In this work, we developed a trajectory-centric methodology for training reasoning models on complete problem-solving workflows. Our approach combines a synthetic data generation pipeline that creates realistic analytical challenges across quantitative domains with a dual reward framework balancing structural consistency and semantic correctness. We incorporated overthinking prevention mechanisms and adaptive curriculum learning to achieve superior token efficiency while maintaining analytical rigor. The resulting dual reasoning interfaces enable both Agentic and CoT mode. Our evaluation demonstrates that this trajectory dataset curation methodology and training recipe surpasses comparable models while also competing with substantially larger models in LCB, AIME and GPQA D benchmarks. 
The model exhibits adaptive reasoning depth, automatically calibrating complexity to task requirements while avoiding the verbosity issues that plague contemporary reasoning systems. These results suggest that process-centric training represents a  advancement in reasoning model development, with implications extending beyond data analysis to broader problem-solving domains.

\appendix


\begin{thebibliography}{99}


\bibitem{ahmad2025opencodereasoning}
Ahmad, W. U., Narenthiran, S., Majumdar, S., Ficek, A., Jain, S., Huang, J., Noroozi, V., and Ginsburg, B. (2025). Opencodereasoning: Advancing data distillation for competitive coding. arXiv preprint arXiv:2504.01943.

\bibitem{chen2025acereason}
Liu, Z., Yang, Z., Chen, Y., Lee, C., Shoeybi, M., Catanzaro, B., and Ping, W. (2025). AceReason-Nemotron 1.1: Advancing Math and Code Reasoning through SFT and RL Synergy. arXiv preprint arXiv:2506.13284.

\bibitem{guo2025deepseekr1}
Guo, D., Yang, D., Zhang, H., Song, J., Zhang, R., Xu, R., Zhu, Q., Ma, S., Wang, P., Bi, X., et al. (2025). Deepseek-R1: Incentivizing reasoning capability in LLMs via reinforcement learning. arXiv preprint arXiv:2501.12948.

\bibitem{hong2024data}
S. Hong, Y. Lin, B. Liu, B. Wu, D. Li, et al. (2024). Data Interpreter: An LLM agent for data science. arXiv preprint arXiv:2402.18679.

\bibitem{hwang2024selfexplore}
Hyeonbin Hwang, Doyoung Kim, Seungone Kim, Seonghyeon Ye, and Minjoon Seo. 2024. Self-explore: Enhancing mathematical reasoning in language models with fine-grained rewards. In \textit{Findings of the Association for Computational Linguistics: EMNLP 2024}, pages 1444–1466.


\bibitem{qwen-team2025qwq}
Qwen-Team. (2025). QwQ-32B: Embracing the Power of Reinforcement Learning. QwenLM Blog.

\bibitem{uesato2022solving}
Jonathan Uesato, Nate Kushman, Ramana Kumar, Francis Song, Noah Siegel, Lisa Wang, Antonia Creswell, Geoffrey Irving, and Irina Higgins. 2022. Solving math word problems with process-and outcome-based feedback. \textit{arXiv preprint arXiv:2211.14275}.

\bibitem{wei2022chain}
Wei, J., Wang, X., Schuurmans, D., Bosma, M., Xia, F., Chi, E., Le, Q. V., Zhou, D., et al. (2022). Chain-of-thought prompting elicits reasoning in large language models. Advances in neural information processing systems, 35, 24824–24837.

\bibitem{xie2023openagents}
T. Xie, F. Zhou, Z. Cheng, P. Shi, L. Weng, et al. (2023). OpenAgents: An Open Platform for Language Agents in the Wild. arXiv preprint arXiv:2310.10634.

\bibitem{yao2022react}
Shunyu Yao, Jeffrey Zhao, Dian Yu, Nan Du, Izhak Shafran, Karthik R. Narasimhan, and Yuan Cao. 2022. ReAct: Synergizing reasoning and acting in language models. \textit{arXiv preprint arXiv:2210.03629}.

\bibitem{hu2024infiagentdabench}
Hu, X., Zhao, Z., Wei, S., Chai, Z., Ma, Q., Wang, G., Wang, X., Su, J., Xu, J., Zhu, M., Cheng, Y., Yuan, J., Li, J., Kuang, K., Yang, Y., Yang, H., and Wu, F. (2024). InfiAgent-DABench: Evaluating agents on data analysis tasks. arXiv preprint arXiv:2401.05507.


\bibitem{openai2024learningreason}
OpenAI. (2024). Learning to reason with LLMs. Retrieved from https://openai.com/index/learning-to-reason-with-llms/

\bibitem{jain2024livecodebench}
Jain, N., Han, K., Gu, A., Li, W.-D., Yan, F., Zhang, T., Wang, S. I., Solar-Lezama, A., Sen, K., and Stoica, I. (2024). LiveCodeBench: Holistic and contamination free evaluation of large language models for code. arXiv preprint arXiv:2403.07974.

\bibitem{shao2024deepseekmath}
Shao, Z., Wang, P., Zhu, Q., Xu, R., Song, J., Bi, X., Zhang, H., Zhang, M., Li, Y. K., Wu, Y., and Guo, D. (2024). DeepSeekMath: Pushing the limits of mathematical reasoning in open language models. arXiv preprint arXiv:2402.03300.

\bibitem{zhang2024benchmarkingdsa}
Zhang, Y., Jiang, Q., Han, X., Chen, N., Yang, Y., and Ren, K. (2024). Benchmarking data science agents. arXiv preprint arXiv:2402.17168.

\bibitem{su2025expandingrl}
Su, Y., Yu, D., Song, L., Li, J., Mi, H., Tu, Z., Zhang, M., and Yu, D. (2025). Expanding RL with verifiable rewards across diverse domains. arXiv preprint arXiv:2503.23829.

\bibitem{luo2025largelanguage}
Luo, J., Zhang, W., Yuan, Y., Zhao, Y., Yang, J., Wu, B., Chen, B., Qiao, Z., Long, Q., Tu, R., Luo, X., Ju, W., Xiao, Z., Wang, Y., Xiao, M., Liu, C., Yuan, J., Zhang, S., Jin, Y., Zhang, F., Wu, X., Zhao, H., Tao, D., Yu, P. S., and Zhang, M. (2025). Large Language Model Agent: A Survey on Methodology, Applications and Challenges. \textit{arXiv preprint} arXiv:2503.21460.


\bibitem{luo2025deepcoder}
Luo, M., Tan, S., Huang, R., Patel, A., Ariyak, A., Wu, Q., Shi, X., Xin, R., Cai, C., Weber, M., Zhang, C., Li, L. E., Popa, R. A., and Stoica, I. (2025). DeepCoder: A fully open-source 14B coder at O3-mini level. Retrieved from https://www.together.ai/blog/deepcoder

\bibitem{lightman2023letsverify}
Lightman, H., Kosaraju, V., Burda, Y., Edwards, H., Baker, B., Lee, T., Leike, J., Schulman, J., Sutskever, I., and Cobbe, K. (2023). Let’s verify step by step. arXiv preprint arXiv:2305.20050.



\bibitem{achiam2023gpt4}
J. Achiam, S. Adler, S. Agarwal, L. Ahmad, I. Akkaya, et al. (2023). GPT-4 Technical Report. arXiv preprint arXiv:2303.08774.

\bibitem{li2024gettingmorejuice}
Li, J., Zeng, S., Wai, H.-T., Li, C., Garcia, A., and Hong, M. (2024). Getting more juice out of the SFT data: Reward learning from human demonstration improves SFT for LLM alignment. arXiv preprint arXiv:2405.17888.


\bibitem{rajbhandari2019zero}
Rajbhandari, S., Rasley, J., Ruwase, O., and He, Y. (2019). ZeRO: Memory optimizations toward training trillion parameter models. arXiv preprint arXiv:1910.02054.

\bibitem{wang2025hierarchicalreward}
Wang, T., Jiang, Z., He, Z., Tong, S., Yang, W., Zheng, Y., Li, Z., He, Z., and Gong, H. (2025). Towards hierarchical multi-step reward models for enhanced reasoning in large language models. arXiv preprint arXiv:2503.13551.

\bibitem{qwen2024technical}
Qwen Team. (2024). Qwen2.5 technical report. arXiv preprint arXiv:2412.15115.

\bibitem{kwon2023pagedattention}
Kwon, W., Li, Z., Zhuang, S., Sheng, Y., Zheng, L., Yu, C. H., Gonzalez, J. E., Zhang, H., and Stoica, I. (2023). Efficient memory management for large language model serving with PagedAttention. arXiv preprint arXiv:2309.06180.

\bibitem{sui2025stopoverthinking}
Sui, Y., Chuang, Y.-N., Wang, G., Zhang, J., Zhang, T., Yuan, J., Liu, H., Wen, A., Zhong, S., Chen, H., and Hu, X. (2025). Stop Overthinking: A Survey on Efficient Reasoning for Large Language Models. arXiv preprint arXiv:2503.16419.

\bibitem{shrivastava2025samplemore}
Shrivastava, V., Awadallah, A., Balachandran, V., Garg, S., Behl, H., and Papailiopoulos, D. (2025). Sample More to Think Less: Group Filtered Policy Optimization for Concise Reasoning. arXiv preprint arXiv:2508.09726.


\bibitem{rein2024gpqa}
Rein, D., Hou, B. L., Stickland, A. C., Petty, J., Pang, R. Y., Dirani, J., Michael, J., and Bowman, S. R. (2024). GPQA: A Graduate-level Google-proof Q\&A Benchmark. In First Conference on Language Modeling.

\bibitem{2025magistral}
Mistral (2025). Magistral. arXiv preprint arXiv:2506.10910.

\bibitem{abdin2024phi4}
Abdin, M., Aneja, J., Behl, H., Bubeck, S., Eldan, R., Gunasekar, S., Harrison, M., Hewett, R. J., Javaheripi, M., Kauffmann, P., Lee, J. R., Lee, Y. T., Li, Y., Liu, W., Mendes, C. C. T., Nguyen, A., Price, E., de Rosa, G., Saarikivi, O., Salim, A., Shah, S., Wang, X., Ward, R., Wu, Y., Yu, D., Zhang, C., and Zhang, Y. (2024). Phi-4 Technical Report. arXiv preprint arXiv:2412.08905.


\bibitem{qwen-team2024qwen}
Qwen-Team. (2024). Qwen2.5: A Party of Foundation Models!. QwenLM Blog.


\bibitem{qwen-team2024qwencode}
Qwen-Team. (2024). Qwen2.5-Coder Series: Powerful, Diverse, Practical. QwenLM Blog.


\end{thebibliography}
\end{document}